%% file: neurips_2026.tex
\documentclass{article}

\usepackage[main,preprint]{neurips_2026}

\usepackage[utf8]{inputenc} 
\usepackage[T1]{fontenc}    

\usepackage{graphicx}
\usepackage{booktabs}
\usepackage{wrapfig}
\usepackage{multirow} 
\usepackage[table]{xcolor}
\usepackage[misc]{ifsym}

\usepackage{hyperref}       
\usepackage{url}            
\usepackage{booktabs}       
\usepackage{amsfonts}       
\usepackage{nicefrac}       
\usepackage{microtype}      
\usepackage{xcolor}         

\usepackage{amsmath}
\usepackage{tabularx}

\title{DriveReward: A Comprehensive Dataset and Generative Vision-Language Reward Model for Autonomous Driving}

\author{%
  \parbox{\linewidth}{\centering
    Qimao Chen\textsuperscript{1,2,*}, 
    Fang Li\textsuperscript{2,*}, 
    Yuechen Luo\textsuperscript{1,2,*}, 
    Zehan Zhang\textsuperscript{2,\textdaggerdbl}, 
    Haiyang Sun\textsuperscript{2}, 
    Fangzhen Li\textsuperscript{2}, 
    Bing Wang\textsuperscript{2}, 
    Guang Chen\textsuperscript{2}, 
    Yang Ji\textsuperscript{2}, 
    Jiong Deng\textsuperscript{2}, 
    Hongwei Xie\textsuperscript{2}, 
    \\
    Hangjun Ye\textsuperscript{2,\Letter}, 
    Long Chen\textsuperscript{2}, 
    Yi Zhang\textsuperscript{1,\Letter}
    \\
    \vspace{0.8em} 
    \mdseries 
    \textsuperscript{1}Tsinghua University \quad
    \textsuperscript{2}Xiaomi EV
    \\
    cqm24@mails.tsinghua.edu.cn
  }
}

\begin{document}

\maketitle

\begingroup
\renewcommand\thefootnote{} 
\footnotetext{* Equal contribution.}
\footnotetext{\textdaggerdbl\ Project Lead.}
\footnotetext{\textsuperscript{\Letter} Corresponding authors.}
\endgroup
\vspace{-10pt}
\begin{figure*}[h!]
\centering
\includegraphics[width=0.9\linewidth]{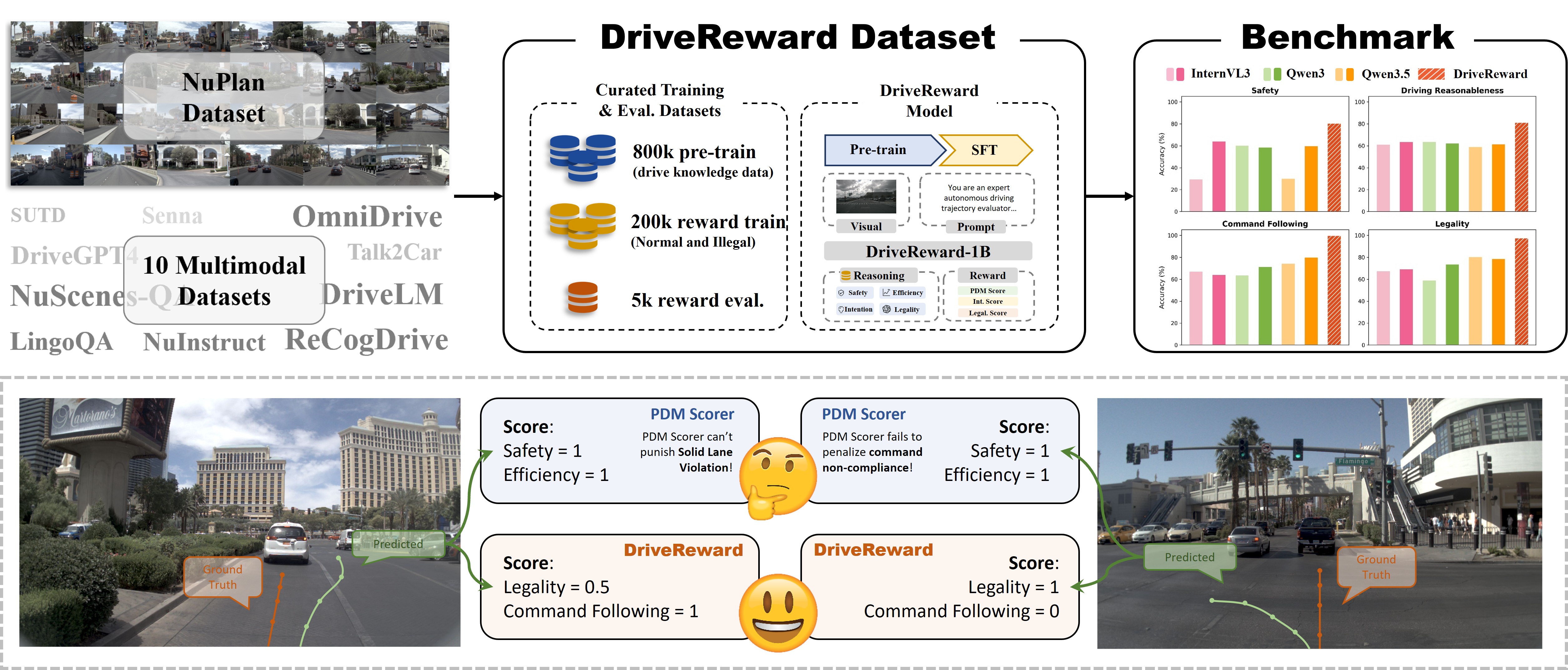}
\label{fig:main}
\end{figure*}
\vspace{-15pt}

\begin{abstract}
  Reward models play a pivotal role in reinforcement learning (RL) and multi-modal trajectory selection for autonomous driving. However, acquiring such rewards typically relies on hand-crafted rule-based objectives or perception ground truth, which hinders generalization for data-scaling. While Vision-Language Models (VLMs) have demonstrated feasibility as reward models in other domains, their effectiveness in driving tasks remains underexplored. In this work, we bridge this gap by (1) introducing DriveReward, a reasoning trajectory evaluation dataset rigorously labeled via temporally-grounded visual guidance, and augmented with counterfactual driving behaviors., (2) alongside a specialized Vision-Language Reward Model.
  To address the scarcity of failure cases in conventional datasets, we propose a counterfactual data annotation scheme to construct cases encompassing diverse driving styles and erroneous behaviors. Evaluations on our proposed benchmark reveal that even leading open-source and proprietary VLMs fail to excel across all tasks, highlighting significant room for improvement in existing models.
  Building on these findings, we subsequently tailor a specialized 1B reward model that outperforms larger VLMs on task-specific reward alignment. Finally, we validate our reward model's effectiveness by integrating it into 
RL finetuning and multi-modal trajectory scoring across multiple baselines, achieving performance comparable to rule-based reward calculations in both open-loop and closed-loop evaluation.
\end{abstract}

\section{Introduction}
\label{sec:intro}

End-to-end planning has recently witnessed rapid advancements in autonomous driving\cite{sun2025sparsedrive, adathinkdrive, fu2025orion, jiang2023vad, luo2025mtrdrive}. Recent methods\cite{li2025hydra, li2024hydra, liao2025diffusiondrive, xing2025goalflow, li2025generalized} focus on predicting higher-quality trajectories. Literature has shown that generating multiple diverse trajectories better captures the multi-modal nature of real-world driving and can improve system performance. In this line of work, the key challenge lies in evaluating these multi-modal trajectories and selecting the optimal one as the final output, which directly affects the safety and efficiency of the system\cite{li2025generalized, li2024hydra, liao2025diffusiondrive, DOR, chen2026vilta}.
Furthermore, in the emerging domain of Vision-Language-Action (VLA) models for autonomous driving\cite{adathinkdrive, li2025recogdrive, zhou2025autovla}, trajectory evaluation is equally critical during reinforcement learning (RL) fine-tuning. When optimizing a policy model's trajectory generation capabilities via RL, the generated trajectories must be accurately scored to provide reliable reward signals. The quality of these reward signals directly determines the efficacy of the fine-tuning process.

\begin{figure}[h!]
    \centering
    \includegraphics[width=1\linewidth]{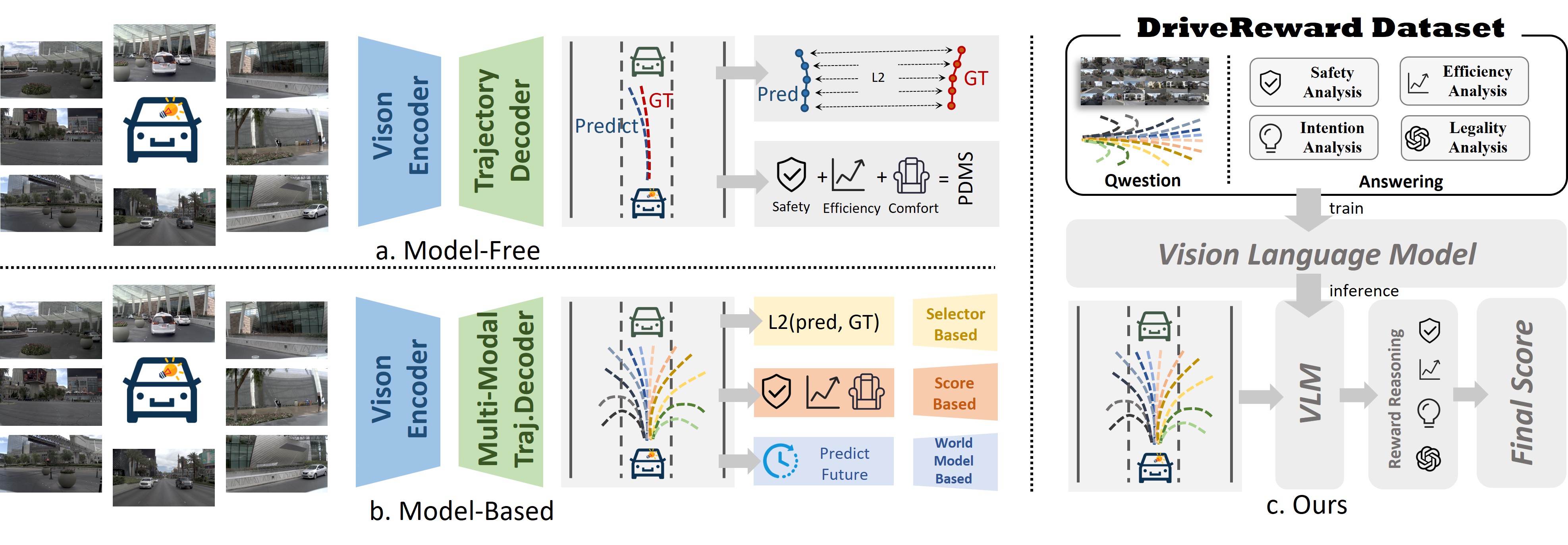}
    \caption{Comparison of trajectory evaluation paradigms. (a) Model-free methods: rule-based scoring using metrics like MSE or PDMS. (b) Model-based methods: trajectory selection via trained selector heads, score heads, or world models. (c) Ours: constructs the multi-dimensional DriveReward dataset and trains a large reward model via a two-stage process for holistic trajectory evaluation.}
    \label{fig:rewardmodelintro}
\end{figure}

\noindent Existing evaluation methods for multi-modal trajectories generally fall into two categories: model-free and model-based approaches (Fig.~\ref{fig:rewardmodelintro} (a) and (b)). Model-free methods \cite{dauner2024navsim, jia2024bench2drive} rely on hand-crafted heuristics from explicit perception outputs. This makes them highly susceptible to perception noise and cascading errors. Furthermore, their strict dependence on dense, expensive annotations severely bottlenecks data scalability. In contrast, model-based methods are typically trained jointly with the planner. Early approaches \cite{liao2025diffusiondrive, sun2025sparsedrive, li2024hydra, li2025generalized} evaluate trajectories using only current perceptual representations, ignoring future scene evolution. Recent works \cite{li2025end, sun2025minddrive} address this by introducing lightweight world models to forecast future states. Nevertheless, these methods remain optimized for dataset-specific proxy metrics (e.g., MSE, PDMS\cite{dauner2024navsim}) and lack high-level semantic reasoning, making it difficult to evaluate complex rules like red light violations.

\noindent To overcome these limitations, we propose a reward model that leverages the powerful visual and semantic reasoning capabilities of Vision-Language Models (VLMs), as shown in Fig~\ref{fig:rewardmodelintro} (c). Using an open-source VLM as the backbone, our model evaluates trajectories by comprehensively understanding the holistic driving context, encompassing road layouts, traffic semantics, and safety constraints. To facilitate the training and evaluation of this model, we construct a large-scale dataset comprising reward-oriented Visual Question Answering (VQA) pairs. Each sample probes the efficiency, safety, and rationality of a specific trajectory within its visual context. Grounded in Large Model annotations, this dataset provides robust and explicit supervision signals for the reward model. Within our dataset, we explicitly introduce two supplementary trajectory evaluation dimensions, effectively addressing the inherent limitations of the PDM score in specific edge-case scenarios.
In summary, we make two main contributions to this work:
\begin{itemize}
    \item \textbf{DriveReward Dataset.} We introduce a novel dataset and benchmark specifically constructed to train and evaluate the trajectory scoring capabilities of Vision-Language Models (VLMs). During the dataset curation, we enriched the data with counterfactual driving behaviors and proposed two novel trajectory evaluation dimensions.
    Furthermore, we devised a fine-grained annotation scheme to explicitly facilitate the temporal understanding of the VLMs employed for annotation.

    \item \textbf{DriveReward Model.} We propose DriveReward-1B, a comprehensive reward model capable of scoring multi-modal trajectories across multiple dimensions.
    Extensive experiments demonstrate the broad utility of DriveReward-1B. Empirical results validate that our model effectively enhances the performance of existing end-to-end planners through RL finetuning and test-time trajectory selection.
\end{itemize}

\section{Related Works}
Recent advancements in end-to-end autonomous driving increasingly rely on multi-modal trajectory generation and VLA models~\cite{sun2025sparsedrive, elf_vla}, making robust trajectory evaluation critical for both inference-time selection and RL alignment. However, existing evaluators are severely bottlenecked by their reliance on explicit perception annotations~\cite{li2025recogdrive, li2024hydra}. Furthermore, while VLMs have shown promise as general-purpose reward models~\cite{yu2025self}, their application in autonomous driving remains largely unexplored. Existing driving reward models, such as Gen-Drive~\cite{gendrive}, primarily rely on preference optimization, struggling to encapsulate the explicit, well-defined metrics required for safe navigation. To bridge these gaps, we propose adapting VLMs as unified, perception-free reward models.
A comprehensive review of related literature is provided in App.~\ref{related_work}.

\section{DriveReward Dataset And Benchmark}
\label{sec:dataset}
To robustly train and evaluate a general reward model for autonomous driving, a diverse dataset encompassing both nominal behaviors and anomalous driving events across varied scenes is indispensable. While several large-scale driving datasets have been publicly released recently (e.g., nuScenes\cite{qian2024nuscenes}, NAVSIM\cite{dauner2024navsim}, Bench2Drive\cite{jia2024bench2drive}), they are predominantly composed of safe, nominal driving scenarios. Consequently, they fail to adequately expose models to the critical long-tail of complex, real-world traffic violations and hazards. Although such data is sufficient for training planners via imitation learning, it remains fundamentally suboptimal for training reward models, which require a strong discriminative capability to accurately assess diverse and potentially unsafe trajectory proposals. To address this critical imbalance, we introduce the DriveReward dataset and benchmark. Our data engine comprises two core components: first, a novel negative-example data augmentation pipeline  that systematically synthesizes long-tail infraction scenarios to broaden the state space coverage; second, a comprehensive, multi-dimensional Question-Answering (QA) pairs built upon these diverse scenarios, designed to facilitate nuanced, fine-grained analysis and evaluation of multi-modal trajectories.

\subsection{Automated Labeling Pipeline}\label{sec:alp}

\begin{figure}[h!]
    \centering
    \includegraphics[width=1\linewidth]{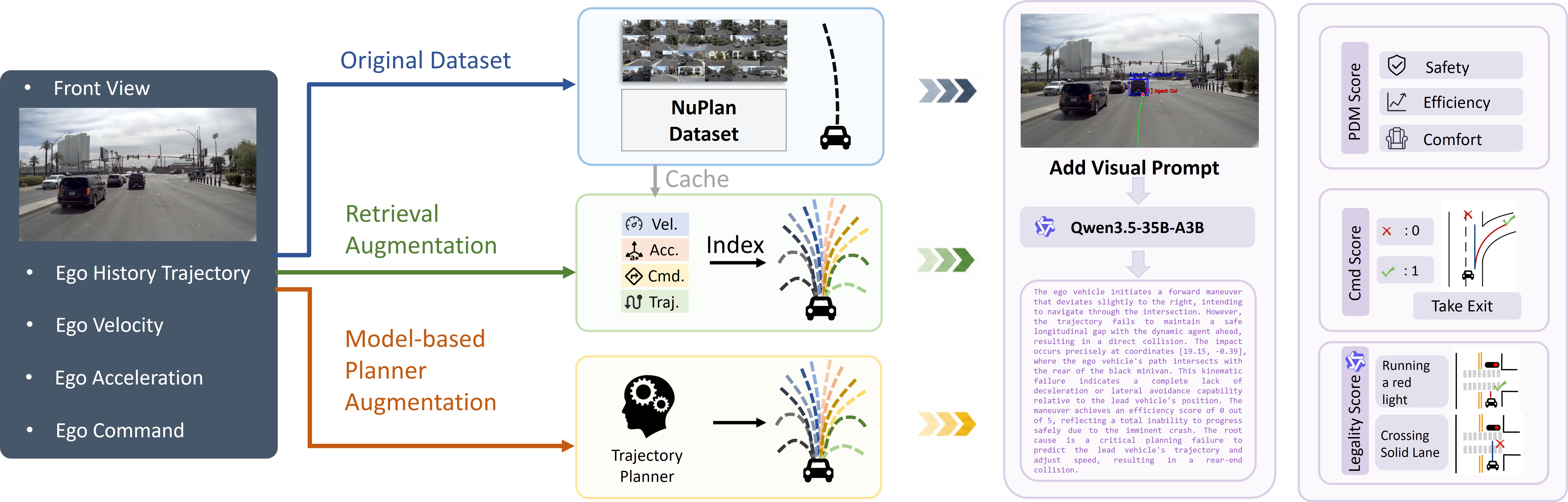}
    \caption{Automated Labeling Pipeline. For a given front-view image and ego state, current-frame trajectories are sourced from the original dataset, ego-state-based retrieval, and planner outputs. These diverse trajectory candidates are then systematically evaluated based on PDMS, command adherence, and legal compliance.}
    \label{fig:pipeline}
\end{figure}

\noindent \textbf{Data Source} We aggregate driving samples from NAVSIM~\cite{dauner2024navsim}\footnote{This work masks use of the NAVSIM dataset. The authors of this work confirm that the use of the above dataset in this work is strictly limited to academic research purposes and does not involve any commercial activities.} train set, a subset of nuPlan\cite{nuplan}.

\noindent \textbf{Automated Labeling.} The overall pipeline is illustrated in Fig.~\ref{fig:pipeline}. We systematically automate the trajectory annotation process through a two-phase approach. Initially, for each driving frame, we construct a diverse pool of trajectory candidates from three sources: the original dataset ground-truths, multi-modal trajectories retrieved from an offline kinematic bank, and planner-generated trajectories aimed at mitigating the human-algorithm domain gap. Following candidate generation, we annotate each trajectory with multi-dimensional reward labels. First, the PDM-Closed simulator computes standard metrics to derive the PDMS. Second, rule-based calculations evaluate the Command Following score. Finally, we prompt the Qwen3.5-35B-A3B~\cite{qwen3.5} large model with the front-view images, trajectories, and the aforementioned scores to perform complex semantic reasoning. 
\textbf{To explicitly facilitate the model's temporal understanding during annotation, we overlay precise visual prompts onto the front-view images. }Specifically, we project the planned trajectory and highlight critical waypoints indicating off-road deviations or collisions, alongside the bounding boxes of the involved surrounding vehicles or road boundaries. Since these visual elements are deterministically computed via rules, they provide reliable spatial-temporal grounding, thereby ensuring the high quality of Qwen3.5's labeling results.
More Details can be found in App.~\ref{details_auto_label}. Consequently, our dataset provides comprehensive trajectory evaluation annotations across seven distinct dimensions. Specifically, the metrics for No Collisions (\textbf{NC}), Drivable Area Compliance (\textbf{DAC}), Ego Progress (\textbf{EP}), Time-to-Collision (\textbf{TTC}), and Comfort (\textbf{C}) are computed strictly following the NAVSIM evaluation protocol. Meanwhile, the formulations for our two newly proposed scores are defined as follows:

    \paragraph{Command Following (CF):} While strict adherence to navigation commands is imperative, existing open-source datasets often lack fine-grained routing annotations (e.g., highway merges or ramp ingress). Consequently, our evaluation is conducted at a coarse-grained maneuver level, verifying fundamental intents such as turning left, turning right, and moving forward. To illustrate the critical necessity of this metric, consider a diverging road scenario where the ground-truth (GT) intent requires merging right to enter a side road. If a predicted trajectory safely but incorrectly proceeds straight ahead without changing lanes, existing evaluation protocols would still assign it a misleadingly high score, as the trajectory remains collision-free and legally compliant despite completely ignoring the navigational objective. To explicitly address such discrepancies, we quantify command adherence by measuring the lateral deviation between the endpoints of the predicted trajectory and the GT trajectory. A binary score is assigned: the trajectory receives a 1 if this deviation falls within a pre-defined threshold, and a 0 if it exceeds it, indicating a fundamental navigation failure.

    \paragraph{Legality (LG):} This metric quantifies trajectory compliance with established traffic rules, focusing primarily on red light violations and solid lane crossings. Rather than relying on rigid geometric heuristics or explicit map annotations, we directly employ the Qwen3.5-35B-A3B large vision-language model to assess these infractions. Given the visual context and the predicted trajectory, Qwen3.5 performs complex semantic reasoning to explicitly output both a textual explanation of the vehicle's behavior and a hierarchical legality score. The scoring logic is rigorously tiered to reflect the varying safety criticalities of these behaviors: if the model identifies a red-light running violation, it assigns a severe penalty score of 0; alternatively, if no red-light violation occurs but the model detects a solid line crossing, it assigns a partial score of 0.5. A trajectory receives a perfect score of 1 only if Qwen3.5 determines it is fully compliant with both rules.

\subsection{DriveReward Benchmark}
Following the construction of our training dataset, we establish DriveReward-Bench based on the NAVSIM test set, consisting of 5k randomly selected samples. This benchmark comprehensively evaluates the reward model's capacity for trajectory understanding and assessment given the current visual observation. While our DriveReward model predicts seven distinct reward metrics, we deliberately exclude Time-To-Collision (TTC) and Comfort (C) from this specific benchmark. As continuous variables, TTC can be adequately proxied by the discrete NC metric, and the Comfort score lacks evaluative significance since it is predominantly 1 across the dataset. Consequently, the evaluation framework is primarily structured around five discrete, boundary-critical dimensions: NC, DAC, Command Following (CF), Legality (LG) and Efficiency (EP).

\section{DriveReward Model}
We propose DriveReward-1B, a novel vision-language reward model tailored for autonomous driving, built upon the InternVL3 \cite{zhu2025internvl3} framework. To ensure real-time deployment feasibility and computational efficiency, we adopt the lightweight InternVL3-1B as our foundational backbone. In the remainder of this section, we elaborate on our proposed approach across two primary dimensions: the structural architecture and the training methodology.

\subsection{Architecture}
\begin{figure*}[t!]
\centering
\includegraphics[width=0.9\linewidth]{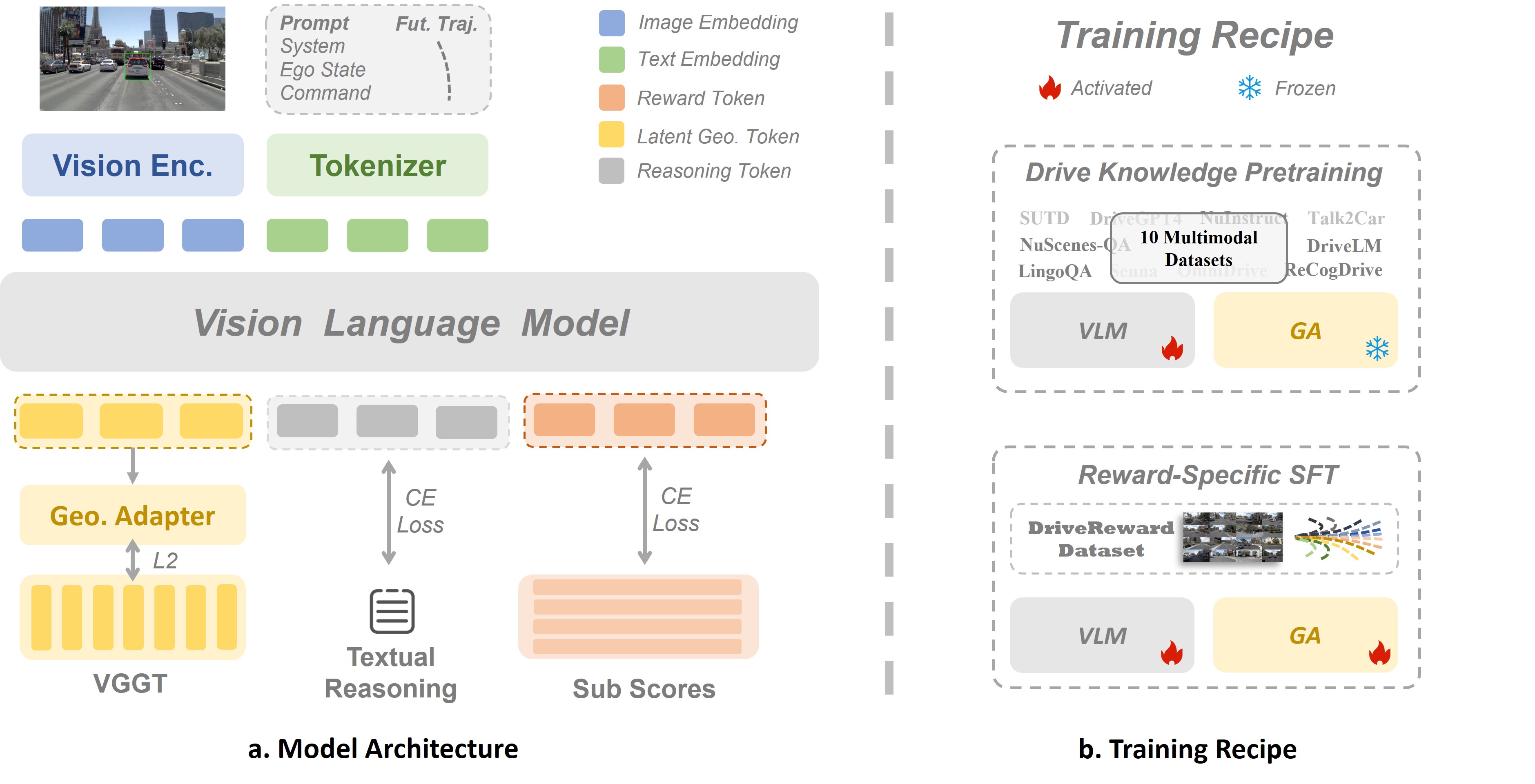}
\caption{DriveReward model and training recipe. (a) DriveReward Model Architecture. Inputs comprise current front-view image and instructions containing ego-state information. The network generates reward tokens, reasoning tokens (supervised by the DriveReward dataset) and geometry latent tokens (supervised via VGGT latents). (b) Two-Stage Training. Initial pre-training builds a foundational understanding of driving scenes, followed by SFT to develop comprehensive multi-dimensional trajectory assessment skills. GA denotes Geo Adapter.}
\label{fig:reward_model}
\end{figure*}

Our DriveReward model architecture is based on InternVL3-1B and enhanced by incorporating a geometry forcing module, as shown in \ref{fig:reward_model} (a). Overall, the model takes current front-view image and textual instructions as inputs. It outputs reward reasoning Chain-of-Thought (CoT) and geometry latent CoT for spatial geometric alignment. The specific formulations of these inputs and outputs are detailed below.

\noindent \textbf{Inputs.} 
At any given time step $t$, the system receives a multimodal query tuple $\mathbf{Q}_t = (I_t, L)$, comprising the front-view camera image $I_t \in \mathbb{R}^{H \times W \times 3}$ and text input $L$, which includes the navigation instruction $l$ (e.g., \textit{``Turn left''}), the ego-vehicle state vector $s_{\text{ego}} \in \mathbb{R}^{d_s}$ (velocity, acceleration), and the trajectory to evaluate $h_{\text{fut}} \in \mathbb{R}^{T \times 3}$.

\noindent \textbf{Geometry Grounding.}
To endow the reward model with robust spatial geometric comprehension—such as accurately assessing whether a trajectory violates drivable area boundaries—we introduce a novel geometry grounding mechanism. Specifically, the reward model is explicitly designed to output an auxiliary geometric feature representation, denoted as $f_{geo} \in \mathbb{R}^{k\times d}$. This feature is subsequently processed by a Geometry Adapter, parameterized as a Multi-Layer Perceptron (MLP), which projects it into the latent space of a pre-trained geometric foundation model, VGGT\cite{wang2025vggt}.  By computing the $L_2$ loss between the projected features and the dense VGGT representations, we establish a direct supervision signal for feature alignment. Through this geometric grounding, the reward model effectively internalizes complex spatial constraints without requiring dense 3D annotations.

\noindent \textbf{Reward Generation.} To empower the Large Reward Model with rigorous trajectory assessment and interpretability, we formulate reward generation entirely as an end-to-end next-token prediction task, completely bypassing the need for auxiliary MLP regression heads. Given the multimodal context, the model autoregressively generates a structured text sequence comprising two distinct segments. First, it produces a comprehensive semantic reasoning of the trajectory's behavior, explicitly encapsulated within \texttt{<think>} and \texttt{</think>} tags. Subsequently, it outputs a robust set of numerical scores enclosed in \texttt{<answer>} and \texttt{</answer>} tags. These quantitative metrics span seven critical driving dimensions: Non-Collision (NC), Drivable Area Compliance (DAC), Time-To-Collision (TTC), Ego Progress (EP), Comfort, Command Following (CF), and Legality. This unified generative process is directly supervised by the reasoning and scoring ground truths provided by the DriveReward Dataset.

\subsection{Training Recipe}\label{training_recipe}
To effectively harness the spatiotemporal features of video data and elicit precise evaluations, we employ a two-stage training paradigm, as shown in Fig~\ref{fig:reward_model} (b). In the first stage, the model undergoes pre-training on a large-scale autonomous driving QA dataset, which endows it with a generalized, foundational understanding of complex driving scenes. In the second stage, we conduct Supervised Fine-Tuning (SFT) specifically on our proposed DriveReward Dataset. This crucial phase aligns the model to deeply analyze historical video frames and accurately formulate reward signals for candidate trajectories.

\noindent \textbf{Stage 1: Domain-Specific Pre-training.} To successfully bridge the domain gap between general-purpose Vision-Language Models (VLMs) and the specialized demands of autonomous driving, our initial training stage focuses on comprehensive domain adaptation. 
Similar to ReCogDrive\cite{li2025recogdrive}, the model is pretrained on a massive corpus aggregating 10 complementary datasets to inject critical world knowledge and driving-specific reasoning into the model. 
These datasets supply essential planning-oriented annotations, action trajectories, and counterfactual reasoning, explicitly linking 3D perceptual scenarios with language-guided decision-making. Details can be found in App.~\ref{training_details}.

\noindent \textbf{Stage 2: Task-Specific Supervised Fine-Tuning.} Building upon the general driving comprehension acquired in the first stage, the second stage performs task-specific supervised fine-tuning (SFT) tailored for trajectory evaluation. We fine-tune the model exclusively on our proposed reward question-answering dataset. 
By conditioning the model on the front-view visual contexts and meticulously designed system prompts {(shown in Fig.~\ref{fig:vis})}, this SFT phase explicitly equips the network with the specialized capabilities necessary for multi-modal trajectory evaluation and accurate reward generation.
It is crucial to note that the visual prompts mentioned in Sec.~\ref{sec:alp} are employed exclusively to assist the annotation process. The actual training of the reward model strictly utilizes the original, unmodified images.

\section{Experiments}
\label{sec:exp}

\subsection{Experimental Setup}

We freeze the reward model's vision backbone during training to preserve its pre-trained representational capacity and exclusively fine-tune the LLM layers.
More details can be found in App.~\ref{training_details}.

\subsection{DriveReward Benchmark}

\begin{table}[h!]
\centering

\caption{\textbf{DriveReward Bench.} Performance comparison of our reward model (DriveReward-1B) with existing zero-shot and task-specific methods. The discrete metrics (Safety, DAC, CF, and Legality) are evaluated using accuracy, whereas the continuous metric (EP) is assessed via Mean Absolute Error (MAE).}

\resizebox{0.7\textwidth}{!}{
\begin{tabular}{lc|ccccc}
\toprule
\textbf{Models} & \textbf{Param} & \textbf{Safety$\uparrow$} & \textbf{DAC$\uparrow$} & \textbf{CF$\uparrow$} & \textbf{Legality$\uparrow$} & \textbf{EP-MAE$\downarrow$} \\
\hline
\multicolumn{7}{c}{\texttt{\texttt{Zero-Shot}}} \\
\hline
\multirow{2}{*}{{InternVL3\cite{zhu2025internvl3}}} 
 & 2B & 29.6 & 61.4 & 67.2 & 67.5 & 0.62 \\
 & 8B & 64.2 & 63.7 & 64.2 & 69.3 & 0.58 \\
 \midrule
\multirow{2}{*}{{Qwen3\cite{yang2025qwen3}}} 
 & 4B & 60.5 & 63.9 & 63.8 & 59.3 & 0.55 \\
 & 8B & 58.8 & 62.4 & 71.4 & 73.8 & 0.55 \\
 \midrule
 \multirow{2}{*}{{Qwen3.5\cite{qwen35blog}}} 
 & 2B & 30.3 & 59.3 & 74.4 & 80.6 & 0.57 \\
 & 4B & 59.9 & 61.5 & 80.1 & 78.8 & 0.60 \\
\hline
\multicolumn{7}{c}{\texttt{Task-Specific}} \\
\hline
InternVL3-1B & 1B & 72.2 & 75.6 & 95.5 & 92.2 & 0.34 \\
\rowcolor{gray!30} DriveReward  & 1B & \textbf{80.6} & \textbf{81.2} & \textbf{99.9} & \textbf{97.6} & \textbf{0.23} \\
\bottomrule
\end{tabular}%
}
\label{tab:drivereward_bench}
\end{table}

\noindent To validate the efficacy of our proposed evaluation framework, we benchmark several state-of-the-art open-source models against our DriveReward model. Specifically, we evaluate various parameter configurations of the InternVL3, Qwen3, and Qwen3.5 families. To establish a strong baseline, we also perform Supervised Fine-Tuning (SFT) on the base InternVL3-1B model using our curated DriveReward dataset. The quantitative results are summarized in Tab.~\ref{tab:drivereward_bench}. As observed, the zero-shot trajectory evaluation capabilities of the open-source models exhibit a strong positive correlation with their parameter scales; larger models consistently yield superior assessment accuracy. Furthermore, following SFT on the DriveReward dataset, the baseline InternVL3-1B model achieves a substantial improvement in overall accuracy. This significant margin demonstrates that our dataset effectively endows VLMs with comprehensive, multi-dimensional trajectory evaluation skills.  Finally, our fully equipped DriveReward model achieves consistent improvements across all evaluated metrics when compared to the SFT-adapted InternVL3-1B.
 This notable gain clearly indicates that our tailored reward model architecture, combined with the two-stage training recipe, drastically enhances the VLM's capacity for nuanced and accurate trajectory assessment.

\subsection{Reward Signal for RL}\label{reward_signal}

\begin{table}[h!]
\centering
\footnotesize
\caption{\textbf{Open-Loop Evaluation.} Performance evaluation of driving policies trained via RL in NAVSIM-V1. We first apply SFT to the base model. Building upon this SFT checkpoint, we conduct RL optimization using our proposed DriveReward-1B (RM-1B).}
\label{tab:rl_train}
\begin{tabularx}{\textwidth}{c|c| >{\centering\arraybackslash}X >{\centering\arraybackslash}X >{\centering\arraybackslash}X >{\centering\arraybackslash}X >{\centering\arraybackslash}X | >{\centering\arraybackslash}X}
\toprule
\textbf{Base~Model} & \textbf{Method} & \textbf{NC}$\uparrow$ & \textbf{DAC}$\uparrow$ & \textbf{TTC}$\uparrow$ & \textbf{C}$\uparrow$ & \textbf{EP}$\uparrow$ & \textbf{PDMS}$\uparrow$ \\
\midrule
\multirow{2}{*}{InternVL3-2B} & SFT & 97.5 &89.7 &\textbf{93.0} &100.0 &74.9 &80.8 \\
 & \cellcolor{gray!30}RL-RM (1B) & \cellcolor{gray!30}\textbf{97.5} &\cellcolor{gray!30}\textbf{91.2} &\cellcolor{gray!30}92.8 &\cellcolor{gray!30}\textbf{100.0} &\cellcolor{gray!30}\textbf{76.6 }& \cellcolor{gray!30}\textbf{82.2} \\
 \midrule
 \multirow{2}{*}{InternVL3-8B} & SFT & 98.3 &93.7 &94.8 &100.0 &79.4 &85.6 \\
 & \cellcolor{gray!30}RL-RM (1B)& \cellcolor{gray!30}\textbf{98.4} &\cellcolor{gray!30}\textbf{94.8} &\cellcolor{gray!30}\textbf{94.9} &\cellcolor{gray!30}\textbf{100.0} &\cellcolor{gray!30}\textbf{80.9} & \cellcolor{gray!30}\textbf{86.8} \\
\midrule
\multirow{2}{*}{DiffusionDriveV2} & SFT & 97.2 &97.0 &92.2 &\textbf{100.0} &86.8 &89.1 \\
 & \cellcolor{gray!30}RL-RM (1B)& \cellcolor{gray!30}\textbf{98.2}& \cellcolor{gray!30}\textbf{97.6} & \cellcolor{gray!30}\textbf{94.4} & \cellcolor{gray!30}99.8 & \cellcolor{gray!30}\textbf{87.2} & \cellcolor{gray!30}\textbf{90.7}\\
\bottomrule
\end{tabularx}
\end{table}

\paragraph{Open-loop Evaluation.}
To validate the efficacy of our proposed reward model for Reinforcement Learning (RL) training, we conduct RL fine-tuning on three distinct base policies: InternVL3-2B, InternVL3-8B and DiffusionDriveV2~\cite{zou2025diffusiondrivev2}. Specifically, we directly use the predicted PDMS from DriveReward-1B as reward signal. The experimental results in NAVSIM-V1 are summarized in Table~\ref{tab:rl_train}. As observed, utilizing our DriveReward model as the guiding reward signal yields consistent improvements over the SFT baselines, demonstrating its capability to provide effective and reliable supervisory signals. 

\begin{table}[h!]
  \centering
  \footnotesize
  \caption{\textbf{Reward Model as RL Enhancer.} The best results for each base model are highlighted in \textbf{bold}. To ensure a fair comparison, all reported results are based on the final training checkpoints.}
  \label{tab:reward_results}
  \begin{tabularx}{\textwidth}{c | c | >{\centering\arraybackslash}X >{\centering\arraybackslash}X >{\centering\arraybackslash}X >{\centering\arraybackslash}X >{\centering\arraybackslash}X | >{\centering\arraybackslash}X}
    \toprule
    \textbf{Base~Model} & \textbf{Method} & \textbf{NC$\uparrow$} & \textbf{DAC$\uparrow$} & \textbf{TTC$\uparrow$} & \textbf{C$\uparrow$} & \textbf{EP$\uparrow$} & \textbf{PDMS$\uparrow$} \\
    \midrule
    \multirow{2}{*}{InternVL3-2B} 
    & Rule-PDMS & 96.9 & 91.3 & 91.5 & 100.0 & \textbf{80.3} & 83.3 \\
    & \cellcolor{gray!30} Rule-PDMS+RM CF\&LG & \cellcolor{gray!30}\textbf{97.4} & \cellcolor{gray!30}\textbf{91.3} & \cellcolor{gray!30}\textbf{92.4} & \cellcolor{gray!30}\textbf{100.0} & \cellcolor{gray!30}79.5 & \cellcolor{gray!30}\textbf{83.3} \\
    \midrule
    \multirow{2}{*}{InternVL3-8B} 
    & Rule-PDMS & 97.9 & 94.8 & 93.6 & 100.0 & \textbf{84.7} & 87.9 \\
    & \cellcolor{gray!30} Rule-PDMS+RM CF\&LG& \cellcolor{gray!30}\textbf{98.1} & \cellcolor{gray!30}\textbf{95.6} & \cellcolor{gray!30}\textbf{94.1} & \cellcolor{gray!30}\textbf{100.0} & \cellcolor{gray!30}84.5 & \cellcolor{gray!30}\textbf{88.4} \\
    \midrule
    \multirow{2}{*}{AdaThinkDrive-8B} 
    & Rule-PDMS & 98.2 & 97.1 & 93.8 & 100.0 & \textbf{85.1} & 89.3 \\
    & \cellcolor{gray!30} Rule-PDMS+RM CF\&LG & \cellcolor{gray!30}\textbf{98.6} & \cellcolor{gray!30}\textbf{97.4} & \cellcolor{gray!30}\textbf{95.4} & \cellcolor{gray!30}\textbf{100.0} & \cellcolor{gray!30}84.5 & \cellcolor{gray!30}\textbf{89.9} \\
    \bottomrule
  \end{tabularx}
\end{table}

To further validate the effectiveness of the Reward Model in RL, we employ it as an enhancer for rule-based rewards. Specifically, during RL training, the composite reward for a given sample is formulated by augmenting the heuristically computed PDMS with a weighted sum of the Command Following (CF) and Legality (LG) scores predicted by our reward model (details shown in App.~\ref{training_details}). As demonstrated in Tab.~\ref{tab:reward_results}, this augmentation strategy effectively boosts the safety performance across various baseline models, albeit with a marginal compromise in driving efficiency.

\paragraph{Closed-loop Evaluation}

\begin{table}[htbp]
  \centering
  \footnotesize
  \caption{\textbf{Closed-loop Evaluation.} Quantitative results comparing our RL reward configurations against existing baseline methods in Bench2Drive.}
  \label{tab:b2d}
  \begin{tabularx}{\textwidth}{l | >{\centering\arraybackslash}X >{\centering\arraybackslash}X >{\centering\arraybackslash}X >{\centering\arraybackslash}X}
    \toprule
    \textbf{Method} & \textbf{Driving~Score}$\uparrow$ & \textbf{Success~Rate}$\uparrow$ & \textbf{Efficiency}$\uparrow$ & \textbf{Comfort}$\uparrow$ \\
    \midrule
    AD-MLP~\cite{zhai2023ADMLP} & 18.1 & 0.0 & 48.5 & 22.6 \\
    UniAD-T.~\cite{hu2023planning} & 40.7 & 13.2 & 123.9 & \textbf{47.0} \\
    UniAD-B.~\cite{hu2023planning} & 45.8 & 16.4 & 129.2 & 43.6 \\
    VAD~\cite{jiang2023vad}   & 42.4 & 15.0 & \textbf{157.9} & 46.0 \\
    \midrule
    Base SFT & 40.2 & 14.5 & 122.1 & 42.9 \\
    \rowcolor{gray!30} RL (RM-Predicted PDMS) & 51.4 & 20.8 & 124.0 & 19.2 \\
    RL (Rule-Based PDMS) & 55.0 & 25.0 & 124.4 & 14.8 \\
     \rowcolor{gray!30} RL (Rule-PDMS + RM CF\&LG) & \textbf{57.1} & \textbf{27.2} & 128.3 & 11.4 \\
    \bottomrule
  \end{tabularx}
\end{table}

We evaluate the zero-shot generalization capability of the trained DriveReward-1B model by deploying it as the reward signal for the RL training of a policy model on the Bench2Drive~\cite{jia2024bench2drive} dataset. Bench2Drive is a comprehensive benchmark built upon the CARLA~\cite{dosovitskiy2017carla} simulator. Specifically, we first perform supervised fine-tuning (SFT) on an InternVL3-8B model using this dataset to establish a baseline policy. Subsequently, we apply RL to optimize this SFT-adapted model under three distinct reward formulations: the PDMS directly predicted by DriveReward-1B, the standard rule-based PDMS, and a composite reward integrating the rule-based PDMS with the CF and LG scores predicted by DriveReward-1B.

As demonstrated in Tab.~\ref{tab:b2d}, employing the PDMS predicted by our DriveReward-1B reward model for RL training substantially enhances the performance of the baseline SFT model (from DS 40.2 to 51.4). Despite being exclusively trained on the nuPlan dataset, our reward model exhibits excellent zero-shot transferability to the Bench2Drive benchmark, validating its strong generalization capabilities. Understandably, a performance gap persists between the model-predicted PDMS and the rule-computed ground-truth PDMS. However, augmenting the ground-truth PDMS with the Command Following (CF) and Legality (LG) sub-scores additionally predicted by DriveReward-1B further amplifies the effectiveness of RL optimization (from DS 55.0 to 57.1). This underscores the exceptional versatility of our reward model: for datasets lacking the infrastructure for rule-based PDMS computation, the model's direct predictions serve as a robust alternative; conversely, for environments where rule-based PDMS is available, integrating our model's supplemental CF and LG predictions via weighting still yields commendable performance gains.

\subsection{Test-time Trajectory Selection}

To validate the multi-modal trajectory evaluation capabilities of our DriveReward model, we conduct test-time trajectory selection experiments on AdaThinkDrive. Results are shown in Tab.~\ref{tab:selection}. In this context, the 'Best-of-N' setting serves as the theoretical upper bound for selection performance. It is computed by executing all $N$ generated trajectory proposals in the simulator and employing an oracle selector to pick the candidate with the highest PDMS score.
  Specifically, we set $N=4$ for AdaThinkDrive (obtained via four independent inference passes). 
 As reported in Tab.~\ref{tab:selection}, the Best-of-N oracle outperforms the vanilla baseline by 3.0\%. Our DriveReward model successfully captures a portion of this potential, delivering a 0.22\% improvement over the baseline.
 Admittedly, the performance gain achieved by the reward model is relatively marginal, indicating that a distinct gap to the theoretical upper bound still persists.
 These quantitative results demonstrate that while DriveReward effectively enhances baseline planning performance through robust trajectory selection, fully closing the gap to the theoretical upper bound remains an open challenge for future exploration.

\subsection{Ablation Studies}

\begin{table*}[t!]
    \centering
    \begin{minipage}{0.55\textwidth}
    \centering
    \caption{\textbf{Reward model as Scorer model.} Performance verifier on multi-modal trajectory selection.}
\label{tab:selection}
\resizebox{\textwidth}{!}{
\begin{tabular}{c|c|ccccc|c}
\toprule
\textbf{Model} & \textbf{Policy} & \textbf{NC}$\uparrow$ & \textbf{DAC}$\uparrow$ & \textbf{TTC}$\uparrow$ & \textbf{CF}$\uparrow$ & \textbf{EP}$\uparrow$ & \textbf{PDMS}$\uparrow$ \\
\midrule
\multirow{3}{*}{AdaThinkDrive\cite{adathinkdrive}} & Original & 98.4 & 97.8 & 95.2 & 100 & 84.4 & 90.3 \\
 & Best of N & 99.1 &98.8  &97.2  &100  &87.9 & 93.0(\textcolor{orange!70!black}{+2.7}) \\
 & Reward Model & 98.0 & 97.5 & 94.1 & 100 & 87.1 & 90.5(\textcolor{green!60!black}{+0.2}) \\
\bottomrule
\end{tabular}
}
\end{minipage}
\hfill
\begin{minipage}{0.42\textwidth}
\centering
\label{ablation_table_models}
\caption{\textbf{Ablation evaluation on DriveReward.}}
\resizebox{\textwidth}{!}{
\begin{tabular}{c|ccccc}
\toprule
\textbf{Ablation} & \textbf{NC}$\uparrow$ & \textbf{DAC}$\uparrow$ & \textbf{CF}$\uparrow$ &\textbf{LG}$\uparrow$ &\textbf{EP-MAE}$\downarrow$  \\
\midrule
\rowcolor{gray!30}DriveReward-1B & \textbf{80.58} & \textbf{81.24} & 99.86 & \textbf{97.64} & \textbf{0.2308} \\

 w/o pre-training & 80.52 & 81.06 & 99.84 & 97.48 &0.2355  \\

w/o reasoning CoT & 71.26 & 73.60 & \textbf{100.00} & - & 0.4161 \\
w/o 3d adapter & 79.04 & 80.88 & 99.86 & 97.20 & 0.2518 \\

\bottomrule
\end{tabular}
}    
\end{minipage}
\end{table*}

To validate the individual contributions and effectiveness of the components within our proposed framework, we conduct comprehensive ablation studies, with quantitative results summarized in Tab.~\ref{ablation_table_models}. Specifically, we investigate three key dimensions: (1) the two-stage training recipe, (2) the trajectory reasoning CoT, and (3) the DriveReward model architecture. Detailed analyses and findings for each aspect are discussed below.

\noindent \textbf{Impact of Training Recipe.} We ablate our two-stage training strategy to quantify its contribution. Although directly fine-tuning (w/o pre-training) the base model establishes a strong baseline, preceding it with domain-specific pre-training on autonomous driving data yields consistent further gains. This underscores the useful role of foundational domain adaptation in maximizing the model's evaluative capabilities.

\noindent \textbf{Impact of Reasoning CoT.} To isolate the impact of the reasoning Chain-of-Thought (CoT), we conduct an ablation study by removing it from the DriveReward dataset. Experimental results (w/o reasoning CoT) reveal that omitting the reasoning CoT significantly degrades the Reward Model's performance. This demonstrates that the reasoning CoT generated by Qwen3.5 is highly effective in enabling the model to comprehensively understand and evaluate the given trajectory.

\noindent \textbf{Impact of Model Architecture.} To validate the effectiveness of the proposed Geometry Adapter, we conduct an ablation study by removing this module from the architecture. Experimental results (w/o 3d adapter) demonstrate that incorporating the Geometry Adapter consistently improves performance across all evaluated metrics. This indicates that explicit spatial geometric supervision significantly enhances the model's trajectory evaluation capabilities.

\section{Limitations}

\noindent \textbf{Dataset Constraints.} Both our proposed dataset and the baseline architecture are currently restricted to single front-view camera inputs, omitting complementary sensory modalities such as LiDAR or multi-view camera systems. Integrating these rich, multi-dimensional data sources would significantly enhance the model's spatial awareness and holistic environmental understanding. Future work will focus on extending the temporal receptive field to process longer historical contexts and incorporating multi-sensor fusion strategies, thereby paving the way for a more generalized and robust reward model suitable for complex real-world deployments. 

\noindent \textbf{Discrepancy with Real-World Deployment.} A final limitation stems from the inherent domain gap between existing open-source datasets and the operational requirements of actual autonomous driving systems. Current public datasets predominantly consist of passive human driving demonstrations, often omitting the complex, multi-modal conditions inherent to real-world navigation. Most notably, they frequently lack fine-grained routing instructions or high-level strategic commands (e.g., GPS-guided lane changes or highway exits), which are indispensable for practical, goal-oriented deployment. Bridging this gap between offline demonstration logs and the interactive, instruction-driven nature of real-world driving remains a critical challenge for future research.

\section{Conclusion}
In this work, we introduce DriveReward dataset and benchmark, designed for training and evaluating generalist vision-language models on multi-modal trajectory assessment. To address this challenge, we propose DriveReward-1B, a dedicated reward model engineered for precise trajectory scoring. Extensive experiments reveal that DriveReward-1B achieves state-of-the-art overall accuracy on DriveReward-Bench across multiple open-source and task-specific baselines, demonstrating that targeted, domain-specific reward supervision enables lightweight models to substantially outperform much larger general-purpose VLMs. Furthermore, DriveReward-1B exhibits commendable efficacy in RL finetuning and test-time multi-modal trajectory selection, highlighting its robust discriminative capability and immense potential for practical deployment in real-world autonomous driving systems.



%
%

\bibliographystyle{unsrt}
\bibliography{mybib}

\newpage
\input{appendix}

\end{document}

%% file: appendix.tex
\section*{Appendix}

In the Appendix, we provide more comprehensive algorithmic details and experimental results. Specifically, Section~\ref{related_work} provides a review of the related literature; Section~\ref{details_auto_label} elaborates on the automated data annotation pipeline; Section~\ref{data_vis} offers qualitative visualizations and further analyses of the curated dataset and benchmark; and Section~\ref{training_details} details the training hyperparameters, hardware configuration and evaluation details.

\subsection{Related Work}\label{related_work}

\paragraph{End-to-End Autonomous Driving.} 
End-to-end (E2E) autonomous driving aims to map raw sensor inputs directly to planning trajectories, effectively streamlining the traditional modular pipeline. Early E2E methods\cite{hu2023planning, jiang2023vad} primarily focus on predicting a single deterministic trajectory. However, recognizing the inherent uncertainty of real-world driving environments, recent works\cite{sun2025sparsedrive, liao2025diffusiondrive, xing2025goalflow, li2024hydra, li2025generalized} have shifted towards generating multi-modal trajectories. This paradigm produces a diverse set of candidate trajectories, subsequently employing a scoring mechanism to select the optimal one for final execution. Concurrently, Vision-Language-Action (VLA)\cite{adathinkdrive, zhou2025autovla, li2025recogdrive, elf_vla} models have emerged as a promising E2E paradigm. By undergoing large-scale pre-training, supervised fine-tuning (SFT), and reinforcement learning (RL), these foundation models acquire robust capabilities for trajectory generation, representing a significant advancement in autonomous driving architectures.

\paragraph{Multi-Modal Trajectory Evaluation.} Trajectory evaluation is essential for ensuring the reliability of autonomous driving systems, serving two critical applications: inference-time trajectory selection and reinforcement learning (RL) fine-tuning. Existing evaluation mechanisms generally fall into two categories: model-free and model-based approaches. Model-free methods directly assess policies using observed data or rule-based scoring, such as L2 tracking loss or weighted metrics derived from perception annotations\cite{adathinkdrive, li2025recogdrive} (e.g., PDMS). Conversely, model-based methods leverage learned environmental dynamics for predictive evaluation. For instance, SparseDrive \cite{sun2025sparsedrive} and WOTE \cite{li2025end} evaluate trajectory candidates using handcrafted functions or BEV features, MindDrive\cite{sun2025minddrive} integrates world models for end-to-end selection, while HydraMDP \cite{li2024hydra} and GTRS \cite{li2025generalized} learn evaluation metrics derived from perception caches. Despite these advancements, both paradigms share a critical bottleneck: their heavy reliance on explicit perception annotations or ground-truth trajectories. This strict dependence severely hinders large-scale data scaling, fails to encapsulate high-level semantic traffic rules, and makes RL alignment on unannotated datasets highly challenging. To bridge these gaps, we propose a Large Reward Model that synthesizes the strengths of both paradigms. By leveraging the extensive prior knowledge of Vision-Language Models (VLMs), our unified model bypasses rigid perception labels, functioning as a superior multi-modal trajectory selector and a dense reward signal provider.

\paragraph{Language Models as Reward model}
Existing approaches to utilizing Vision-Language Models (VLMs) as reward models generally fall into three paradigms. The first paradigm \cite{xiong2025llava} directly prompts the model to output a numerical score or ranking. While straightforward, this method relies heavily on the model's zero-shot instruction-following and comprehension capabilities. The second approach \cite{zang2501internlm, lou2024uncertainty} maps the hidden representations of the VLM to a task-specific regression head (typically a linear layer or MLP) to predict continuous or discrete scores. Although computationally efficient, it fundamentally lacks interpretability. The third paradigm \cite{yu2025self, zhang2502mm} simultaneously trains the model to generate evaluative reasoning (e.g., critiquing question-answer pairs) and outputs a scalar reward via an auxiliary head. This hybrid approach strikes a balance between interpretability and efficiency, albeit demanding specialized data formats and complex training strategies. Meanwhile, VLMs have demonstrated remarkable potential for open-world understanding and semantic reasoning in autonomous driving. 
Gen-Drive~\cite{gendrive} introduces a Reward Model trained via Direct Preference Optimization (DPO) using preference pairs. However, unlike natural language generation, autonomous driving tasks are governed by explicit, well-defined evaluation metrics. Consequently, a reward model derived solely from preference data is insufficient to comprehensively evaluate driving trajectories.
To this end, leveraging VLMs as dedicated reward models remains largely unexplored. Therefore, adapting VLMs for trajectory evaluation and RL fine-tuning presents a highly promising and timely research direction.

\subsection{Details of Auto Labeling}\label{details_auto_label}

\begin{figure}[h!]
    \centering
    \includegraphics[width=1\linewidth]{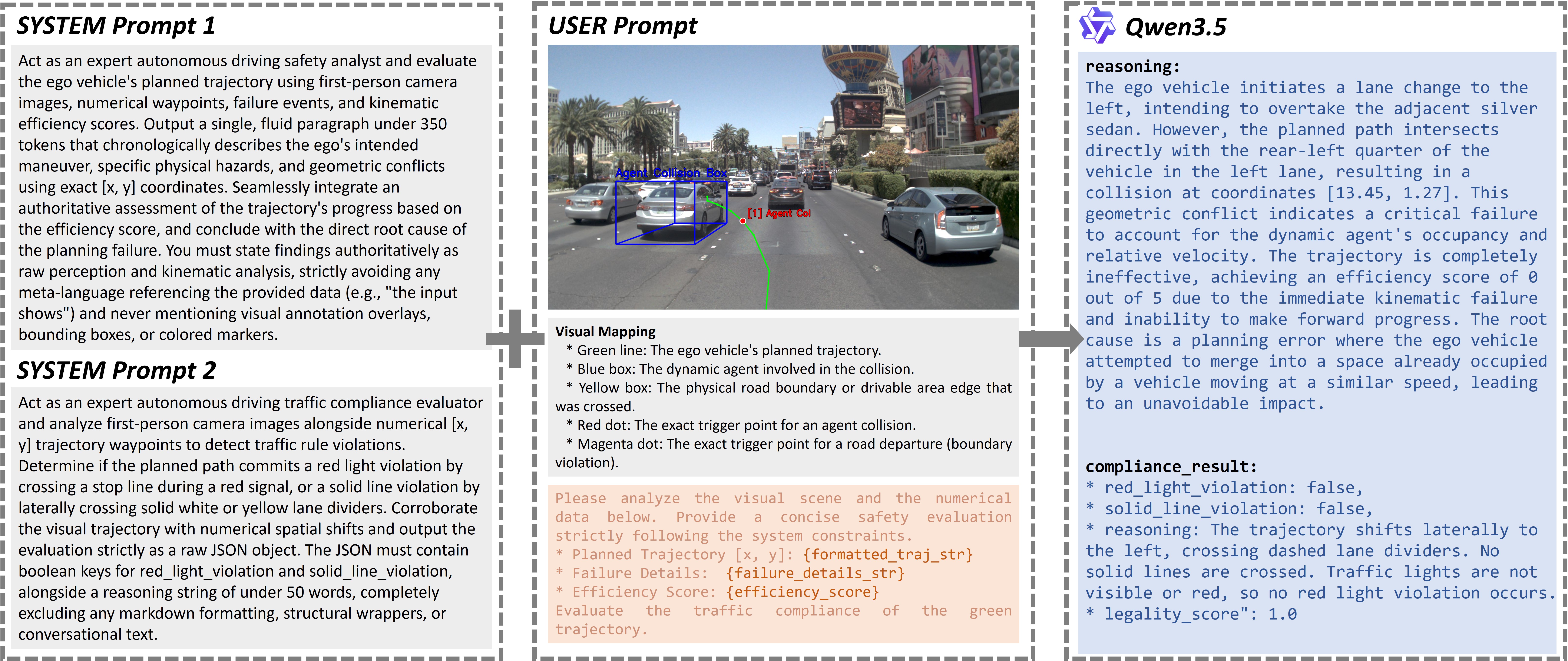}
    \caption{A representative example of reasoning data generation pipeline. System Prompts 1 and 2 target trajectory evaluation reasoning and legality score generation, respectively. The user prompt comprises images with annotated visual prompt of safety hazards and the ego-vehicle's past and current states. Ultimately, the model outputs the step-by-step reasoning for trajectory assessment and a detailed rule compliance analysis.}
    \label{fig:reasoning}
\end{figure}

 As shown in Fig.~\ref{fig:reasoning}, Qwen3.5 executes a dual-task evaluation: it explicitly interprets the rationale behind the assigned PDMS and command scores, while simultaneously assessing the trajectory's adherence to visual traffic rules (e.g., crossing solid lines or running red lights) to yield a final legality score. 
 To facilitate Qwen3.5's temporal understanding during the annotation process, we overlay visual prompts onto the front-view images. Specifically, the green line represents the trajectory under evaluation, red points indicate collision locations with other vehicles, and magenta points denote where the trajectory deviates from the drivable area. Additionally, blue boxes highlight the bounding boxes of the collided vehicles, while yellow boxes demarcate the departed road regions.

\subsection{Dataset Analysis}\label{data_vis}

\begin{figure}[h!]
    \centering
    \includegraphics[width=1\linewidth]{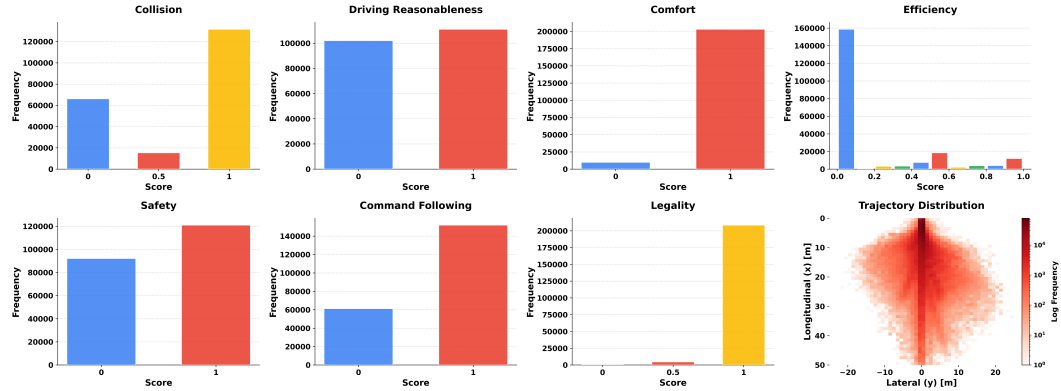}
    \caption{Distributions of sub-metrics and trajectories within the DriveReward dataset.}
    \label{fig:ana}
\end{figure}

\begin{figure}[htb]
	\centering
	\includegraphics[width=\linewidth]{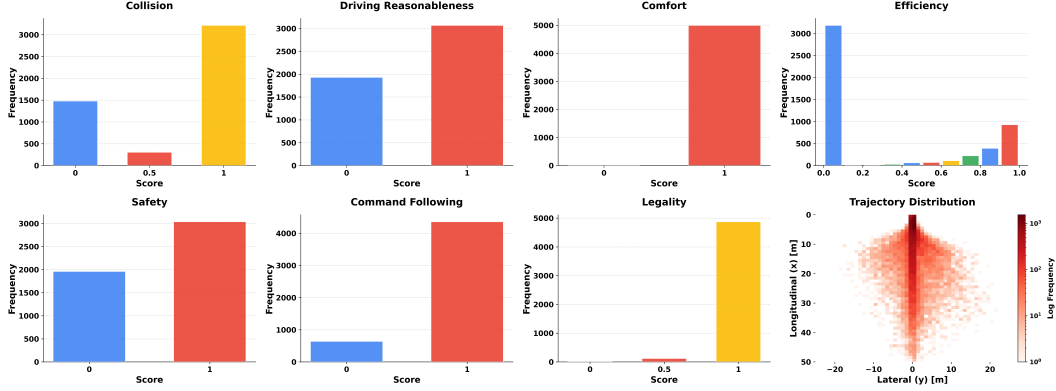}
    \caption{Distributions of sub-metrics and trajectories within the DriveReward Benchmark.}
    \label{fig:bm_ana}
\end{figure}
The distributions of individual sub-scores and trajectories in our proposed DriveReward dataset are detailed in Fig.~\ref{fig:ana}.
In total, we curated 200K samples. Benefiting from our counterfactual augmentation, the proportion of sub-optimal trajectories in the dataset has been substantially enriched. Here, sub-optimal trajectories are strictly defined as those failing to achieve a perfect score of 1 in Collision, Driving Reasonableness, or Safety. Regarding our two newly introduced sub-scores, approximately 30\% of the trajectories receive a Command Following score of 0, indicating that a considerable fraction fails to comply with driving commands. Conversely, the ratio of trajectories with Legality violations remains relatively low, implying that the vast majority successfully avoid running red lights or crossing solid lines.
Figure~\ref{fig:bm_ana} illustrates the data distribution of the 5K samples in the benchmark.

\begin{figure}[h!]
    \centering
    \includegraphics[width=1\linewidth]{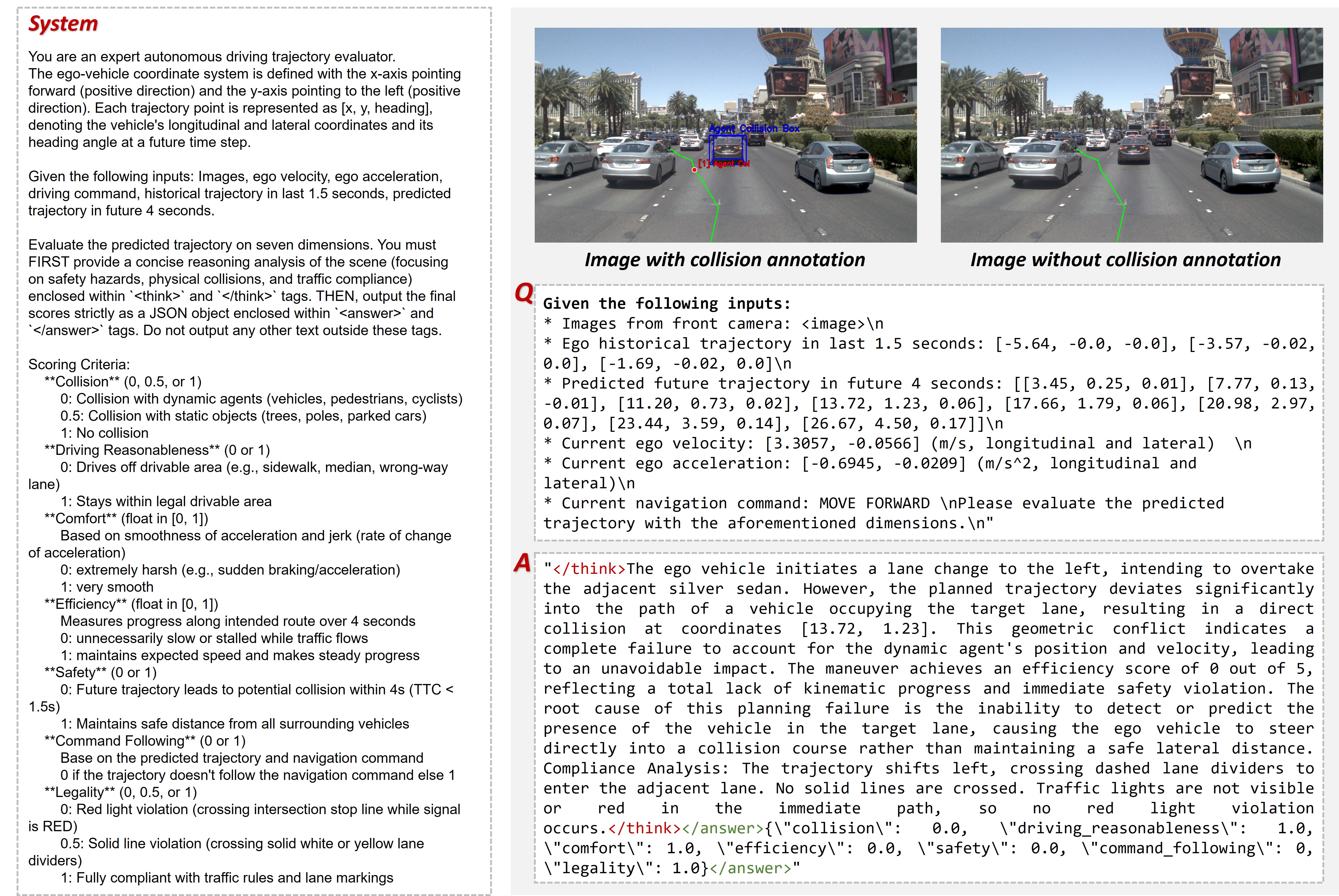}
    \caption{Visualization of two typical samples from the DriveReward dataset. Each sample comprises a 3-frame historical video sequence coupled with textual ego-state information. The corresponding annotations provide numerical metrics across seven distinct evaluation dimensions: Safety, DAC, TTC, EP, Comfort, Command Following, and Legality (note that TTC, EP, and Comfort are omitted in this visualization).}
    \label{fig:vis}
\end{figure}

Fig.~\ref{fig:vis} illustrates a representative sample from the DriveReward dataset. 
In this scenario, while the explicit navigational command dictates "MOVE FORWARD," the predicted trajectory executes a left lane change, ultimately resulting in a collision with a leading vehicle. However, because the ego-vehicle neither crosses a solid lane boundary nor violates a red traffic light, it strictly complies with visual traffic regulations, thereby yielding a perfect Legality score of 1. Accompanying this, the reasoning Chain-of-Thought (CoT) provides a comprehensive semantic analysis of the complex scene, explicitly justifying the rationale behind each individual sub-score.

\subsection{Experimental Details}\label{training_details}

\paragraph{Training Parameters and Hardware Configuration.}
We utilize InternVL3-1B~\cite{zhu2025internvl3} as the foundation model, training across two sequential stages on 32 NVIDIA H20 GPUs, as described in Sec.~\ref{training_recipe}. 
Specifically, the two-stage training proceeds as follows: in the first stage, we conduct full parameter fine-tuning on the 800k domain-specific QA dataset for 6 epochs, setting the learning rate to 4e-5 and gradient accumulation steps to 8. In the second stage, we perform full parameter fine-tuning on the 300k DriveReward dataset for 8 epochs, utilizing a learning rate of 4e-4 and a gradient accumulation step of 1. All training procedures are implemented using the ms-swift~\cite{zhao2024swiftascalablelightweightinfrastructure} framework.

\paragraph{Details of Pretrain Dataset.}
For foundational spatial and multi-modal scene comprehension, we integrate DriveLM\cite{sima2024drivelm}, ReCogDrive\cite{li2025recogdrive}, NuScenes-QA\cite{qian2024nuscenes}, SUTD\cite{SUTD} and Talk2Car\cite{deruyttere-etal-2019-talk2car}, which collectively provide abundant question-answer pairs derived from complex urban environments. To extend this understanding into the temporal domain, we incorporate LingoQA\cite{marcu2024lingoqa} and NuInstruct\cite{ding2024holistic}. The former contributes dense, video-level QA pairs focused on continuous localization and descriptive reasoning, while the latter enforces rigorous spatial-temporal and multi-view reasoning across diverse subtasks.  Finally, to align these visual representations with actual driving execution, we utilize Senna\cite{jiang2024senna}, OminiDrive\cite{wang2025omnidrive} and DriveGPT4\cite{xu2024drivegpt4}. 

\paragraph{Details of Reward Designs in RL.}

In Sec.~\ref{reward_signal}, we compare three distinct variants of reward signals: the rule-based PDMS, the Reward Model-predicted PDMS, and the rule-based PDMS augmented with the CF and LG scores predicted by the Reward Model. In the following, we explicitly detail the mathematical formulations for calculating these three rewards.

PDMS integrates five sub-metrics: No At-Fault Collision (NC), Drivable Area Compliance (DAC), Time-to-Collision (TTC), Comfort (C), and Ego Progress (EP) to produce a comprehensive closed-loop planning score. Its calculation formula is defined as follows:
\begin{equation}\label{equation:pdms}
PDMS = NC \times DAC \times \left( \frac{5\times EP + 5\times TTC + 2\times C}{12} \right),
\end{equation}
For the PDMS predicted by the Reward Model, the overall formulation remains identical Eq.~\ref{equation:pdms}, with the distinction that the constituent sub-scores are directly predicted by the model rather than derived from handcrafted rules. Meanwhile, the calculation for the hybrid reward—integrating the rule-based PDMS with the model-predicted Command Following (CF) and Legality (LG) scores—is formulated as follows:
\begin{equation}\label{equation:cflg}
PDMS_{CFLG} = NC \times DAC \times \left( \frac{5\times EP + 5\times TTC + 2\times C + 2\times CF_{pred} + 2\times LG_{pred}}{16} \right),
\end{equation}
where $CF_{pred}$ and $LG_{pred}$ are Command Following score and Legality score predicted by the reward model, respectively.